\documentclass[9pt,shortpaper,twoside,web]{ieeecolor}
\usepackage{generic}
\usepackage{cite}
\usepackage{multirow}

\usepackage{amsmath,amssymb,amsfonts}
\usepackage{algorithmic}
\usepackage{graphicx}
\usepackage{textcomp}
\usepackage{ulem,url,makecell}
\def\BibTeX{{\rm B\kern-.05em{\sc i\kern-.025em b}\kern-.08em
		T\kern-.1667em\lower.7ex\hbox{E}\kern-.125emX}}
\begin{document}
	\title{Exploring Contextual Relationships for Cervical Abnormal Cell Detection}
	\author{Yixiong Liang, \and Shuo Feng, \and Qing Liu, \and Hulin Kuang, \and Jianfeng Liu, \and Liyan Liao, \and Yun Du,
		\and Jianxin Wang \IEEEmembership{Senior Member, IEEE}
		\thanks{This work was supported in part by the National Key R\&D Program of China (No.2021YFF1201202), the National Natural Science Foundation of China (No.62006249), the Hunan Provincial Science and Technology Innovation Leading Plan (No.2020GK2019), and the Hunan Provincial Natural Science Foundation of China (No.2021JJ40788). }
		\thanks{Y. Liang, S. Feng, Q. Liu, H. Kuang and J. Wang are with the School of Computer Science and Engineering, Central South University, Changsha 410083, China (e-mail: \{yxliang,jxwang\}@csu.edu.cn). }
		\thanks{J. Liu is with School of Automation, Central South University, Changsha 410083, China (e-mail: ljf@csu.edu.cn).}
		\thanks{L. Liao is with the Department of Pathology, The Second Xiangya Hospital, Central South University, Changsha, China.}
		\thanks{Y. Du is with the Department of Cytology, The Fourth Hospital of Hebei Medical University, Shijiazhuang, China.}
	}	
	\maketitle
\begin{abstract}
	Cervical abnormal cell detection is a challenging task as the morphological discrepancies between abnormal and normal cells are usually subtle. To determine whether a cervical cell is normal or abnormal, cytopathologists always take surrounding cells as references to identify its abnormality. To mimic these behaviors, we propose to explore contextual relationships to boost the performance of cervical abnormal cell detection. Specifically, both contextual relationships between cells and cell-to-global images are exploited to enhance features of each region of interest (RoI) proposals. Accordingly, two modules, dubbed as RoI-relationship attention module (RRAM) and global RoI attention module (GRAM), are developed and their combination strategies are also investigated. We establish a strong baseline by using Double-Head Faster R-CNN with feature pyramid network (FPN) and integrate our RRAM and GRAM into it to validate the effectiveness of the proposed modules. Experiments conducted on a large cervical cell detection dataset reveal that the introduction of RRAM and GRAM both achieves better average precision (AP) than the baseline methods. Moreover, when cascading RRAM and GRAM, our method outperforms the state-of-the-art (SOTA) methods. Furthermore, we also show the proposed feature enhancing scheme can facilitate both image-level and smear-level classification. The code and trained models are publicly available at \url{https://github.com/CVIU-CSU/CR4CACD}.
\end{abstract}


\section{Introduction}
\label{sec:introduction}
\IEEEPARstart{C}{ervical} cancer is the fourth most common cause of cancer incidence and mortality among women, with approximately 570,000 confirmed cases and 311,000 deaths worldwide in 2018 \cite{arbyn2020estimates,elakkiya2022cervical}. Nevertheless, cervical cancer is preventable, and early diagnosis is essential to improve the survival rate of cervical cancer. Cytology‐based screening has been the paradigm for the diagnosis and prevention of cervical cancer for more than half a century, which has greatly improved the diagnosis rate of precancerous lesions and cancerous lesions at the cellular level. The cervical cytology is typically diagnosed by cytopathologist on either traditional Pap smears or liquid-based preparation slides \cite{siebers2009comparison} using a microscope to first find out abnormal cells or lesions from tens of thousands of cells, and then give the final smear classification according to the Bethesda System (TBS) \cite{nayar2015bethesda} for reporting cervical cytology. However, manual examination of cervical cytology slides, just like looking for a needle in a haystack, is often tedious, labor-intensive, subjective and prone to error \cite{ma2020pathsrgan}. 
\begin{figure*}[h]
	\centering
	\includegraphics[width=1.0\textwidth]{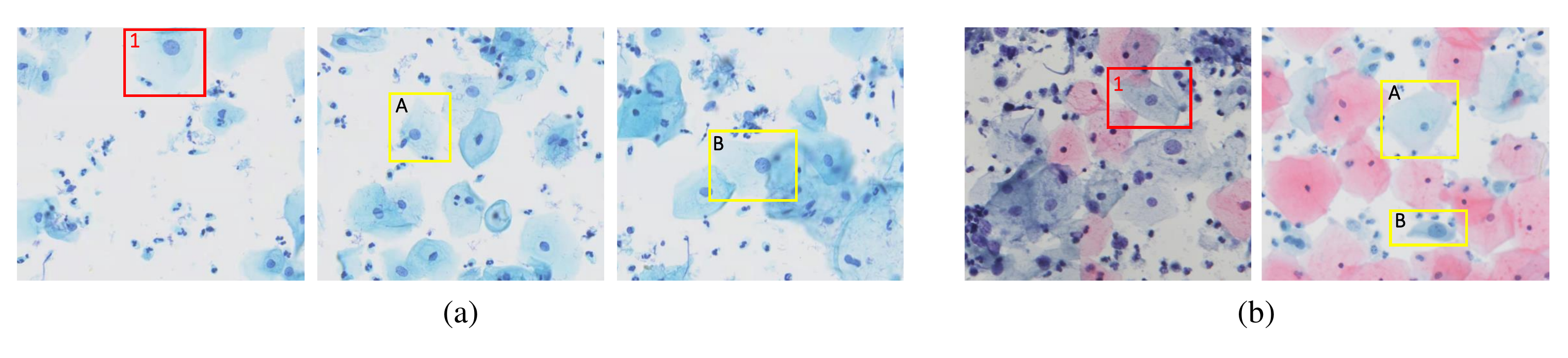}
	\caption{Two examples to illustrate the ambiguous results with different reference cells. The target cell in the red box {\ttfamily{cell 1}} is more likely to be diagnosed as `ASCUS' with the reference {\ttfamily{cell A}}, but as normal with the reference {\ttfamily{cell B}}.}
	\label{img:cell_ref}
\end{figure*}

In order to improve the efficacy and efficiency of cervical cytological screening, the development of automated cervical cytology analysis is in extraordinary demand. There are lots of computer-aided analysis methods to detect the abnormal cells or lesions from cervical cytology images. Most of the early methods follow a two-stage framework, i.e. cell segmentation and abnormality classification \cite{mitra2021cytology,landau2019artificial}. The performance of these methods, however, is often limited by the segmentation accuracy and the hand-crafted feature representation ability \cite{Zhang2017deeppap}. With the overwhelming success in the analysis of natural images, deep learning (DL) has also been applied to cervical cell segmentation \cite{zhou2020deep,song2015accurate,liang2021weakly} and classification \cite{Zhang2017deeppap}. Very recently the DL-based object detection methods \cite{ren2017faster,lin2017feature,lin2018focal,redmon2018yolov3,tian2019fcos} are applied to directly detect the abnormal cells from cervical cytology images in an end-to-end manner and have taken this domain by storm \cite{zhu2021hybrid,lin2021dual,liang2021global,liang2021comparison}. 

Although these modern detection-based methods have achieved appealing performances, they do not take full advantage of the domain knowledge. When detecting an object (e.g. car or cat) in natural images, it is often sufficient to extract the feature from the object area or the region of interest (RoI) to locate and classify the object simultaneously. However, for cervical abnormal cells detection, normal and abnormal cells may have very similar appearances, hence such a local reasoning alone is often not enough. Clinically, to determine whether a cervical cell is normal or abnormal, cytopathologist usually compares it to the surrounding reference cells, then identifies its category. As shown in Fig.~\ref{img:cell_ref}, the cell in the red box ({\ttfamily{cell 1}}) can be identified as abnormal (`ASCUS') when referred to {\ttfamily{cell A}}, but as normal in comparison with {\ttfamily{cell B}}, namely the cervical cell classification can not just rely on the features extracted from the cell patch or the RoI. Existing methods \cite{zhu2021hybrid,lin2021dual,liang2021global,liang2021comparison} often lack the feature interaction between cells, leading to the suboptimal classification performance. 

In this paper, to imitate the behavior of cytopathologists, we propose to explore both the cell relationship and the global image context to boost the performance of cervical abnormal cell detection. Specifically, we propose a novel cascaded RoI feature enhancement scheme based on the attention mechanism \cite{vaswani2017attention,carion2020end}. As shown in Fig. \ref{fig:method}, we introduce the cell relationship attention module and the global attention module into the proposal-based detection framework (e.g. Double-Head Faster R-CNN \cite{wu2020rethinking}) to take advantage of the relational contextual information between cervical cells and the global contextual information respectively to improve the RoI feature representation ability. The main contributions of this work can be summarized as follows:
\begin{itemize}
	\item We propose to simultaneously explore cell-level object-object relationships and the global image context for  cervical abnormal cell detection.	
	\item We devise a cascaded RoI feature enhancement scheme, consisting of an RoI-relationship attention module (RRAM) and a global RoI attention module (GRAM), to enrich the RoI feature representation.	
	\item We validate the effectiveness of our cervical cell detection method on a large cervical cytology image dataset and our method consistently achieves better overall performance comparing to SOTA general and specific detection methods. We also show that our method can facilitate the image-level and smear-level classification. 
\end{itemize}

\section{Related work}\label{related_work}
\subsection{Cervical cytology image analysis}
Since the conventional cytology-based screening, i.e. Papanicolaou test or Pap test, was widely accepted by the medical community in the 1940s, researchers have never stopped their exploration of automated methods for cervical cytology image analysis. The common pipeline usually includes a segmentation step followed by an abnormality classification step \cite{mitra2021cytology,landau2019artificial}. The most recent promising advances include modern object detection-based cervical abnormal cell detection \cite{liang2021global,liang2021comparison,chen2022task} and smear-level classification \cite{pirovano2021computer,lin2021dual,wei2021efficient,jiang2022deep,geng2022}. 

\textbf{Cervical cell segmentation and classification.} Due to the fact that cytological criteria for identifying cervical cell abnormalities are mostly based on the morphological changes of cells and cytoplasm, the segmentation of cervical cell or cellular components (i.e. nuclei and cytoplasm) is regarded as a critical step and numerous cervical cell segmentation works have been devoted \cite{zhou2020deep,song2019segmentation,zhou2019irnet,zhao2022net}. 
Unfortunately, cervical cell segmentation is still a unsolved problem partially due to the large shape and appearance variation between cells, occlusion, poor contrast, etc \cite{Lu2017evaluation}.

The classification of segmented cervical cell patch involves cellular feature extraction and classification algorithms. Popular features in the literature include cell size, shape, color and textural characteristics related to the malignant associated changes described in TBS \cite{nayar2015bethesda}. However, the extraction of these hand-crafted features often depends on the accurate segmentation of cervical cell and once the segmentation error is taken into account, the classification accuracy will significantly decrease \cite{Zhang2017deeppap,liang2021comparison}. As a \textit{de facto} trend, convolutional neural network (CNN) is used to learn the features for cervical cell classification instead of using the hand-crafted ones \cite{Zhang2017deeppap,mitra2021cytology,rahaman2021deepcervix,qin2022multi}. But these methods only aim at the single cell classification and the number of forward pass is proportional to the cell populations (typically about 20,000{\raise.17ex\hbox{$\scriptstyle\sim$}}50,000 cells per slide), making the automated analysis very computationally expensive.

\textbf{Cervical abnormal cell detection.} There has been a surge of interest in improving accuracy and efficiency by virtue of modern object detection techniques. In \cite{zhang2019dccl}, the original Faster R-CNN \cite{ren2017faster} and RetinaNet \cite{lin2018focal} are migrated to detect six kinds of cervical lesion cells while in \cite{zhu2021hybrid}, the native YOLOv3 \cite{redmon2018yolov3} is trained to detect infected and malignment cervical cells. Xiang \textit{et al.} \cite{xiang2020novel} propose a customized YOLOv3 with an additional task-specific classifier to detect 10 subtyping including squamous, glandular and infectious lesions. The Comparison detector \cite{liang2021comparison} is proposed to circumvent the problem of the limited data in cervical abnormal cell detection. Nevertheless, all aforementioned methods only take advantage of the local features, neglecting useful contextual information. 

To employ the contextual information, Liang \textit{et al.} \cite{liang2021global} propose a global context-aware framework to reduce false positive predictions by introducing an extra image-level classification branch. \cite{lin2021dual} present a DP-Net which concatenates both local cellular feature and global image feature for classification. Cao \textit{et al.} \cite{CAO2021102197} add channel attention and spatial attention into Faster R-CNN to boost cell detection performance. Different from these methods, we try to encode both relationships between cells and global contextual information to enrich RoI features based on attention mechanism \cite{vaswani2017attention}.

Concurrent to our work, Chen \textit{et al.} \cite{chen2022task} independently propose a dynamic comparing module (DCM) based on the dynamic filters \cite{sun2021sparse} to imitate clinical manner of cytopathologists. While both methods are conceptually similar, our implementation is based on attention mechanism \cite{vaswani2017attention} and is focused on the interactions between both cells and global image which are ignored in \cite{chen2022task}.

\textbf{Smear-level classification.} Very recently, there are emerging automatic analysis of cervical smear via digitized gigapixel whole slide image (WSI) \cite{jiang2022deep}, which aggregate the cell-level predictions into smear-level diagnosis by either hand-crafted \cite{lin2021dual,pirovano2021computer} or learned strategies \cite{zhu2021hybrid,cheng2021robust,geng2022}. The most straightforward scheme is thresholding on the predicted abnormal probability and the number of abnormal cells \cite{lin2021dual,pirovano2021computer}, which is often sensitive to the inevitable errors of lesion detection due to the large population of cells, resulting in poor specificity. A more sophisticated strategy is to aggregate the engineered features \cite{zhu2021hybrid} or learned features \cite{wei2021efficient,zhou2021hierarchical} of the top-N detected lesions and train a classifier to remedy the cell-level prediction errors \cite{cheng2021robust,wei2021efficient}. Our work tries to improve the accuracy of cell-level predictions and thereby will facilitate the smear-level classification. 
\begin{figure*}[ht]
	\centering
	\includegraphics[width=0.8\textwidth]{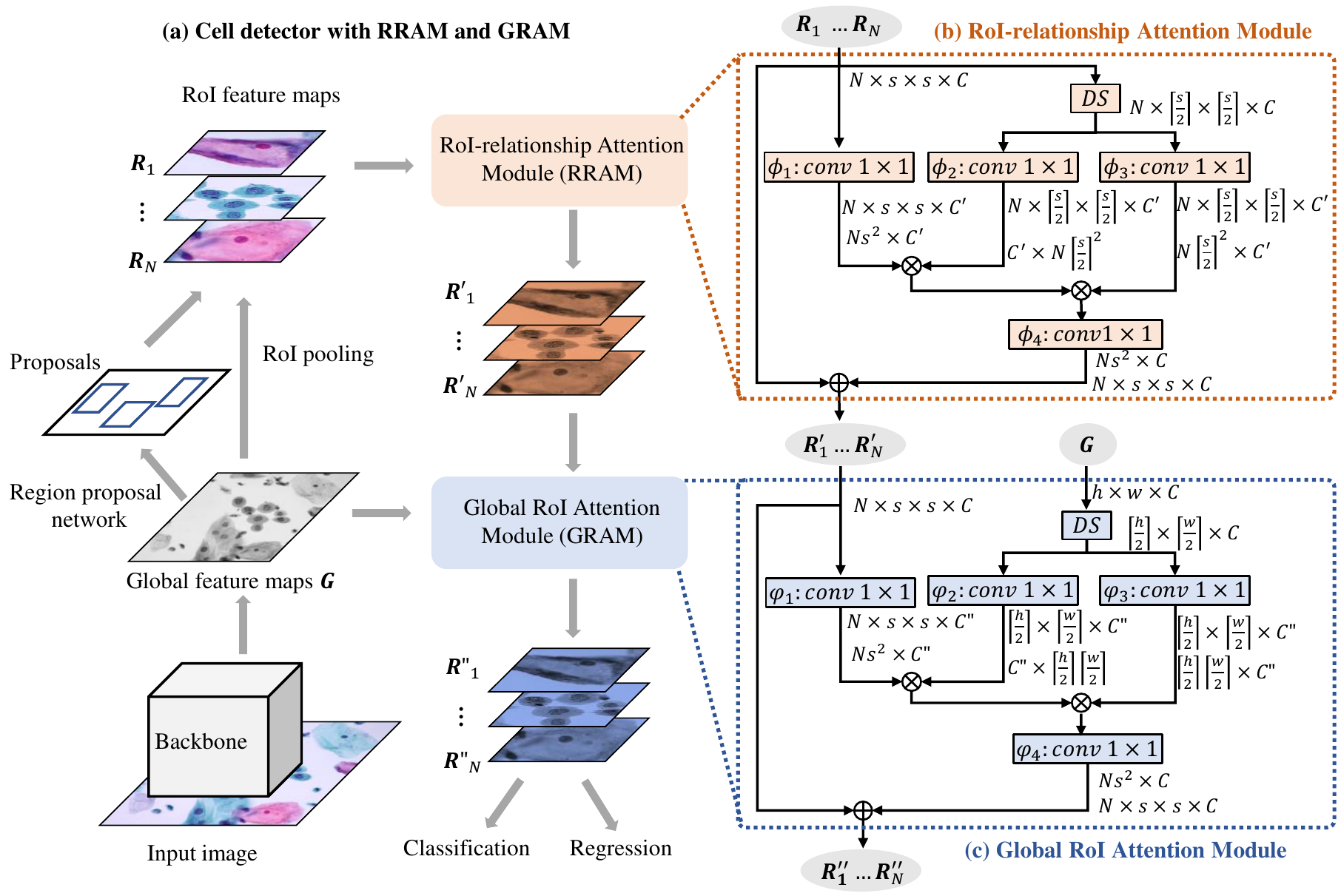}
	\caption{The overview of cell detector with our RRAM and GRAM. It is built on the top of Faster R-CNN equipped with FPN \cite{ren2017faster,lin2017feature}. As shown in (a), with the RoI feature maps $\{\mathbf{R_n}\}_{n=1}^N$ where $\mathbf{R}_n\in \mathbb{R}^{s\times s \times C}$ provided by the region proposal network and RoI pooling layer, our RRAM exploits the relationship both inner the same RoI and across RoIs to enhance $\{\mathbf{R}_n\}_{n=1}^N$, and obtains $\{\mathbf{R}'_n\}_{n=1}^N$. Our GRAM further enhances RoI feature maps via exploiting global contexts from FPN feature maps. (b) and (c) illustrate the details implemented via matrix operations, where $DS$ represents spatial downsampling, $conv 1\times 1$ represents a convolution operator with $1\times 1$ kernel, $\oplus$ represents the element-wise summation and $\otimes$ represent the matrix multiplication.}
	\label{fig:method}
\end{figure*}

\subsection{Object detection with context modeling} 
\textbf{DL-based object detection.}
Existing modern object detectors are mostly categorized by whether having a RoI proposal step (two-stage) \cite{ren2017faster,lin2017feature,dai2016r,wu2020rethinking} or not (one-stage) \cite{redmon2018yolov3,lin2018focal,tian2019fcos,dai2021dynamic}. Most of them are anchor-based, using either anchor boxes \cite{ren2017faster,redmon2018yolov3,lin2018focal,wu2020rethinking} or anchor points \cite{tian2019fcos,zhang2020bridging,dai2021dynamic}. Moreover, they need hand-crafted components like NMS to remove duplicate boxes \cite{sun2021sparse}. In contrast, the Transformer-based DETR \cite{carion2020end} and its variants \cite{zhu2021deformable,zhang2022dino} are fully anchor-free without using hand-designed components and therefore can perform end-to-end optimization.
Generally, one-stage methods are conceptually simpler and much faster, but currently the two-stage detectors have domination in accuracy \cite{qiao2021detectors}. Following \cite{liang2021comparison,chen2022task}, we choose the popular Faster R-CNN \cite{ren2017faster,wu2020rethinking} with feature pyramid network (FPN) \cite{lin2017feature} as candidates of baseline because of its high accuracy and flexibility. 

\textbf{Context modeling in object detection.} Context modeling has long been used as complementary information to facilitate detection. The common contextual information includes local context, global context and relationships between objects. Local context can be integrated by simply enlarging the proposal size \cite{zeng2018crafting,zhu2017couplenet} while global context can be incorporated by either global pooling operation \cite{li2018r,liang2021global,lin2021dual} or learning with recurrent neural networks (RNN) \cite{bell2016inside,li2017attentive}. Closely related to our GRAM, Wang \textit{et al.} \cite{wang2018non} introduce a non-local network to model global contextual information. Nevertheless, in non-local network all pixel-wise features need to interact each other, whereas in our GRAM, only the pixel-wise features in RoI are involved to harvest global contextual information. 

Modeling instance-level relationships in object detection is more challenging. In \cite{chen2017spatial,chen2018iterative}, a RNN-like spatial reasoning and global graph-reasoning is exploited to model the object-object relationship. SIN \cite{liu2018structure} also adopts graph structure to model the relationship among objects and whole scene simultaneously. ReCoR \cite{chen2021recursive} decomposes the contextual relationships into spatial modeling and channel-wise modeling which are progressively modeled in a recursive manner. RelationNet \cite{hu2018relation} and DETR \cite{carion2020end} are the most similar approaches to our RRAM which also exploit the self-attention to model the object relationships. DETR \cite{carion2020end} along its variants (e.g. Deformable DETR \cite{zhu2021deformable}) reasons about the relations of objects and global context to directly predict class and location of objects in parallel, but the introduction of Transformer encoder-decoder architecture \cite{vaswani2017attention} makes it data-hungry. While RelationNet \cite{hu2018relation} treats the RoI features as a whole to compute the object relationships, our RRAM resorts to fine-grained pixel-wise features. 

\section{Method}
In this section, we first present the overview of a two-stage cervical cell detector with the proposed plugged-in RoI-relationship Attention Module (RRAM) and Global RoI Attention Module (GRAM), then introduce the detailed design of RRAM and GRAM.
\subsection{Framework overview}
Fig. \ref{fig:method}(a) illustrates the overview flowchart of the cervical cell detector equipped with the proposed two contextual relationship exploring modules. It conforms to the two-stage object detection paradigm. First, the backbone network extracts convolutional features for the input image. Then a region proposal network (RPN) together with RoI pooling \cite{ren2017faster, he2020mask} are employed to get region proposals and the corresponding RoI feature maps. Subsequently, our RRAM is employed to capture the contextual relationship across cell proposals to enhance RoI feature maps, and GRAM follows to capture global context for further feature enhancement. Finally, these enhanced RoI feature maps are forwarded into the detection head for cell classification and bounding box regression. Concretely, for sake of both accuracy and efficiency, we instantiate our method with the Double-Head Faster R-CNN \cite{wu2020rethinking}, which uses two different heads for classification and regression respectively. The loss function is the same as in the Faster R-CNN \cite{ren2017faster}, including the classification/regression loss of both the RPN and the detection head. Next, we will present the details of our RRAM and GRAM.

\subsection{RoI-relationship attention module (RRAM)}
As we have mentioned before, to accurately identify the abnormal cells, cytopathologists usually not only rely on characteristics about suspected abnormal cells but also refer to typical normal cells near suspected abnormal cells. In other words, both features about suspected abnormal cell itself and the relationship between other cells nearby help the cell identification. To this end, we propose RRAM to explore relationships across cell proposals for RoI feature enhancement.

Supposing that RPN outputs $N$ RoIs and their feature maps provided by RoI pooling layer are denoted as $\{\mathbf{R}_n\}_{n=1}^N$ and of size $s\times s \times C$ where $s$ is the height and width of RoI and $C$ is the number of channels, our RRAM aims to enhance the feature at each position in all RoIs via aggregating the influence from features inside the same RoI as well as other RoIs. Inspired by \cite{vaswani2017attention,wang2018non}, we adopt attention mechanism for feature enhancement. First, our RRAM calculates a normalised feature similarity between feature $\mathbf{r}_{n,i,j}\in \mathbb{R}^{1\times C}$ at location $(i,j)$ in $\mathbf{R}_n$ and feature $\mathbf{r}_{m,k,l}\in \mathbb{R}^{1\times C}$ at location $(k,l)$ in $\mathbf{R}_m$ via:
\begin{equation}
Sim(\mathbf{r}_{n,i,j}, \mathbf{r}_{m,k,l}) = \delta(\frac{\phi_1(\mathbf{r}_{n,i, j})\phi_2(\mathbf{r}_{m,k,l})^T}{\tau})\;,
\label{eq:roi_sim}
\end{equation}
where $m,n\in [1, \cdots, N]$, $i,j,k,l \in [1, \cdots, s]$, and $\delta(\cdot)$ is a softmax function which normalises the similarity into a range of $[0,1]$, and $\tau$ is a scaling factor, and $\phi_{1}(\cdot)$ and $\phi_{2}(\cdot)$ are two linear functions parametrised with $\mathbf{w}_1\in \mathbb{R}^{C\times C'}$ and $\mathbf{w}_2\in \mathbb{R}^{C\times C'}$ respectively, which are used to project the features from $C$-dimensional space into $C'$-dimensional space. We follow \cite{wang2018non, vaswani2017attention} and set $\tau$ to $C'$.

Then our RRAM aggregates all influences from other features on $\mathbf{r}_{n,i,j}$, yielding RoI-relationship influence vector $\mathbf{z}_{n,i,j}$ via:
\begin{equation}
\mathbf{z}_{n,i,j} = \sum_{m,k,l}Sim(\mathbf{r}_{n,i,j}, \mathbf{r}_{m,k,l}) \phi_{3}(\mathbf{r}_{m,k,l})\;,
\label{eq:roi_influence}
\end{equation}
where $\phi_3(\cdot)$ is a linear function mapping the features in $C$-dimensional space into $C'$. Obviously, in Eq.(\ref{eq:roi_influence}), when $m=n$, $Sim(\mathbf{r}_{n,i,j}, \mathbf{r}_{m,k,l})$ measures the relationship between locations $(i,j)$ and $(k,l)$ inside the same RoI $\mathbf{R}_n$. When $m\neq n$, $Sim(\mathbf{r}_{n,i,j}, \mathbf{r}_{m,k,l})$ measures the relationship between two locations across two different RoIs, i.e., $\mathbf{R}_n$ and $\mathbf{R}_m$. However, in Eq.(\ref{eq:roi_influence}), $Sim(\cdot, \cdot)$ has to be calculated $N\cdot s^2$ times, resulting heavy computational cost. To reduce the computational cost, we downsample $\mathbf{r}_{m, k, l}$ and only aggregate influences from locations when $mod(k,2)=0$ and $mod(l,2)=0$. In this way, $Sim(\cdot, \cdot)$ only needs to be calculated $N\cdot \lceil \frac{s}{2} \rceil^2$ times.

Finally, with the aggregated influence vector $\mathbf{z}_{n,i,j}$, our RRAM enhances the feature via
\begin{equation}	
\mathbf{r}'_{n,i,j} = \phi_4(\mathbf{z}_{n,i,j}) + \mathbf{r}_{n,i,j}\;,
\label{eq:roi_enhance}
\end{equation}
where $\phi_4(\cdot)$ is a linear function mapping features in $C'$-dimensional space back to $C$-dimensional space. For each feature in each RoI, we enhance them via our RRAM, then we obtain the enhanced RoI feature maps, denoted by $\{\mathbf{R}'_n\}_{n=1}^N$. 

For simplicity, Eq.(\ref{eq:roi_sim}), Eq.(\ref{eq:roi_influence}) and Eq.(\ref{eq:roi_enhance}) can be implemented via matrix operation:
\begin{equation}
\begin{aligned}
	&		\mathbf{R} = [\mathbf{R}_1, \cdots, \mathbf{R}_N]\;,\\
	&		\mathbf{R}_{\downarrow} = downsample(\mathbf{R})\;,\\
	&	\mathbf{Z} = v^{-1}\left(\phi_4\left( \delta\left (\frac{v(\phi_1(\mathbf{R}))v(\phi_2(\mathbf{R}_{\downarrow}))^T}{\tau}\right)v\left(\phi_3(\mathbf{R}_{\downarrow})\right)\right)\right)\;,\\
	& \mathbf{R}' = \mathbf{Z} + \mathbf{R}\;,
\end{aligned}
\label{eq:roi_enhance_matrix}
\end{equation}
where $\mathbf{R}, \mathbf{Z}$ and $\mathbf{R}'$ are tensors of size $N\times s \times s \times C$ and $\mathbf{R}_{\downarrow}$ is the downsampled tensor of size $N\times \lceil \frac{s}{2} \rceil \times \lceil \frac{s}{2} \rceil \times C$ and $v(\cdot)$ is a tensor reshape operator which reshapes a tensor of size $a_1 \times a_2 \times a_3 \times a_4$ to a matrix of size $a_1a_2a_3\times a_4$ and $v^{-1}(\cdot)$ is an inverse of $v(\cdot)$. Fig. \ref{fig:method}(b) illustrates the implementation details.

\subsection{Global RoI attention module (GRAM)}
Our GRAM is designed to explore global context to enhance the RoI feature maps for better cervical abnormal cell detection. To this end, we reuse feature maps provided by FPN.

Formally, let $\mathbf{G}$ be the feature map of size $h\times w \times C$ from FPN, for each feature $\mathbf{r}'_{n,i,j}$ in RoI-relationship enhanced feature maps $\mathbf{R}_n'$, our GRAM aggregates the influence from entire feature maps $\mathbf{G}$ via:
\begin{equation}
\mathbf{h}_{n, i, j} = \sum_{k,l} \delta(\frac{\psi_1(\mathbf{r}'_{n,i, j})\psi_2(\mathbf{g}_{k,l})^T}{\tau}) \psi_{3}(\mathbf{g}_{k, l})\;,
\end{equation}
where $\mathbf{g}_{k,l}\in\mathbb{R}^{1\times C}$ is the feature vector at location $(k, l)$ in global feature map $\mathbf{G}$, and $\psi_{1}(\cdot), \psi_{2}(\cdot), \psi_{3}(\cdot)$ are the linear functions which projects features from $C-$dimensional space into $C''-$dimensional space. Similar to RRAM, to reduce the computational cost, we downsample $\mathbf{g}_{k,l}$ and only aggregate features when $mod(k, 2)=0$ and $mod(l,2)=0$. With $\mathbf{h}_{n, i, j}$, we then obtain the final enhanced feature via:
\begin{equation}
\mathbf{r}''_{n, i, j} = \psi_{4}(\mathbf{h}_{n, i, j}) + \mathbf{r}'_{n,i,j}\;,
\end{equation}
where $\psi_4(\cdot)$ is a linear function mapping the features in $C''$-dimensional space back to $C$-dimension. Via performing our GRAM on each feature in $\mathbf{R}_n'$ yields the final enhanced feature maps $\mathbf{R}''$. The matrix operation based implementation of our GRAM can be expressed as:
\begin{equation}
\begin{aligned}
	&	\mathbf{G}_{\downarrow} = downsample(\mathbf{G})\;,\\
	&	\mathbf{H} = v^{-1}\left(	\psi_4\left( \delta\left (\frac{v(\psi_1(\mathbf{R}'))v(\psi_2(\mathbf{G}_{\downarrow}))^T}{\tau}\right)v\left(	\psi_3(\mathbf{G}_{\downarrow})\right)\right)\right)\;,\\
	& \mathbf{R}'' = \mathbf{H} + \mathbf{R}'\;,
\end{aligned}
\label{eq:g_enhance_matrix}
\end{equation}
where $\mathbf{G}_{\downarrow}$ is the downsampled tensor of size $\lceil \frac{h}{2} \rceil \times \lceil \frac{w}{2} \rceil \times C$, $\mathbf{H}=[\mathbf{H}_1, \cdots, \mathbf{H}_N]$ and $\mathbf{R}''=[\mathbf{R}''_1, \cdots, \mathbf{R}''_N]$ are tensors of size $N\times s \times s \times C$. The implementation details are illustrated in Fig. \ref{fig:method}(c).

\section{Experiments: Cervical abnormal cell detection}
\subsection{Dataset and experiments settings}

\begin{table*}[tbh!]
\centering
\caption{The detailed class distribution of annotated instances in our CCD dataset.}	
\label{tab:dataset}
\begin{tabular}{l|ccccccccccc}
	\hline
	& Normal & ASCUS & ASCH & LSIL & HSIL & AGC & VAG & MON & DYS & EC & total \\
	\hline
	\texttt{train} & 38,103 & 26,593 & 14,678 & 8,899 & 11,341 & 11,568 & 10,995 & 2,782 & 10,968 & 10,111 & 146,038 \\
	\texttt{val} & 6,302 & 4,421 & 2,517 & 1,528 & 1,942 & 2,128 & 1,819 & 459 & 1,491 & 1,694 & 24,301 \\
	\texttt{test} & 6,414 & 4,371 & 2,564 & 1,513 & 1,857 & 2,062 & 2,014 & 458 & 1,534 & 1,754 & 24,541 \\
	\hline
\end{tabular}
\end{table*}

\begin{table}[tbh!]
\centering
\caption{The detailed scale distribution of annotated boxes in our CCD dataset.}
\label{tab:dataset_size}
\begin{tabular}{l|ccc}
	\hline
	& small & medium & large  \\
	& $(0, 32^2]$ &  $(32^2, 96^2]$ &  $(96^2, +\infty)$ \\
	\hline
	\texttt{train} & 19.5\% & 65.8\% & 14.7\%  \\
	\texttt{val} & 20.6\% & 64.9\% & 14.5\% \\
	\texttt{test} & 19.8\% & 65.5\% & 14.7\% \\
	\hline
\end{tabular}
\end{table}
\textbf{Dataset.} To validate the effectiveness of our method, we collect a cervical cytology images dataset, called \textit{Cervical Cell Detection} (CCD) dataset, consisting of 40,000 images from 1,588 cervical liquid-based cytology smears (1,233 abnormal smears and 355 normal smears) with Papanicolaou stain. Each smear is digitized into a gigapixel WSI via instruments from BioPIC Ltd with 0.1725 $\mu m$/pixel under 20$\times$ magnification, which is then divided into non-overlapping field of view (FoV) images with resolution of $4,096 \times 2,816$. 
Only 20\textasciitilde 30 FoV images per slide are selected for annotation. Each FoV image is first labeled by experienced cytotechnologists via a home-made semi-automatic annotation software and then cross validated by at least another experienced cytopathologists. The final annotations were produced by a consensus of three cytopathologists in case of inconsistent results between two pathologists \cite{cheng2021robust}. 	

Conforming to TBS categories \cite{nayar2015bethesda}, 194,880 object instance bounding boxes are annotated, including  squamous or glandular lesions, infected benign and normal instances. In that the distribution of each class is highly unbalanced, we merge some subtypes according to the high similarity between them. For instance, squamous-cell carcinoma (SCC) is merged into high-grade squamous intraepithelial lesion (HSIL) and all glandular lensions are merged into atypical glandular cells (AGC). Furthermore, due to the utilization of semi-automatic annotation tool, some hard negative instances (Normal) are also annotated to assist the model training, but are not involved during performance evaluation. By this way, all annotated instances are classified into 10 categories, i.e., Normal, atypical squamous cells-undetermined significance (ASCUS), atypical squamous cells-cannot exclude HSIL (ASCH), low-grade squamous intraepithelial lesion (LSIL), HSIL, AGC, vaginalis trichomoniasis (VAG), monilia (MON), dysbacteriosis (DYS) and endocervical cells (EC), etc.

We randomly split the 40,000 images in our CCD dataset into training set (\texttt{train}), validation set (\texttt{val}) and test set (\texttt{test}) with a ratio of 6:1:1. The details of class distribution of annotated instances are listed in Table \ref{tab:dataset}. We compare ours with SOTA methods on the CCD \texttt{test} and report ablations on the CCD \texttt{val}.   It is worthy mentioning that the original size of FoV images ($4,096\times 2,816$) is too big for the network. Following the commonly-used COCO size which has a short side of 800 pixels, we simply resize them into $1,164\times 800$ to reduce the computational burden \cite{liang2021weakly}. After resizing, the detailed size distribution of annotated instances is listed in Table \ref{tab:dataset_size}. 

\textbf{Evaluation metrics.} To conduct quantitative evaluation, we use the standard COCO-style AP metrics and average recall (AR) \cite{lin2014microsoft}. The COCO-style AP metric is the mean over bounding box IoU thresholds ranging from 0.5 to 0.95 with a step of 0.05. The higher the IoU threshold is set, the more accurate the cell location is. AP$_{50}$ and AP$_{75}$ denote AP at fixed 50\% and 75\% IoU threshold. The metrics of AP$_S$, AP$_M$ and AP$_L$ are AP for small ($area \leq 32^2$), medium ($32^2 < area \leq 96^2$) and large ($area > 96^2$) cells, respectively.

\textbf{Implementation details.} We build our cell detector on the top of Double-Head Faster R-CNN \cite{wu2020rethinking} with ResNet-50 \cite{he2016deep} as backbone followed by a FPN \cite{lin2017feature}, which is therefore adopted as baseline. We set double-head for classification and regression, which is 2-layer fully connected network for classification and 2-layer fully convolutional network for regression. 
The training epoch is set to 12 and learning rate is initialized by 0.01 with a decay by a factor of 0.1 after 8 and 11 epochs. We adopt stochastic gradient descent (SGD) optimizer with momentum 0.9 and weight decay 0.0001. Parameters in backbone network ResNet-50 are initialised with the pretrained ImageNet model while the rest are initialised with Kaiming method \cite{he2015delving}. The channel numbers in Eq.(\ref{eq:roi_sim}), Eq.(\ref{eq:roi_influence}) and Eq.(\ref{eq:g_enhance_matrix}) are all set to 128. We set the classification and regression loss weight of the RPN to 1.0, and the classification and regression loss weight of the detection head to 2.0. 

\subsection{Ablation experiments} 
We first run a number of ablations on the CCD \texttt{val} to analyze the proposed modules and discuss in detail next.

\textbf{Downsampling operator selection in RRAM.} In our RRAM, there are several options for the downsampling operators. Here we explore four different downsampling operators: naive subsample with downsampling rate of 2, maxpooling with stride of 2, linear projection with kernel size of $2\times 2$ and stride of 2, and global average pooling (GAP). To explore the influence of different downsampling operators, we conduct experiments and report the performances in Table \ref{tab:DS}. As we can see that, no matter what downsampling operator is applied, our RRAM consistently boosts the cell detection performances. Local downsampling operators, i.e., maxpooling and naive subsampling perform better than global average pooling. The possible reason is that local downsampling operators are able to preserve more spatial details than global average pooling and these spatial details contribute to RoI feature enhancement. Among the local downsampling operators, the naive subsampling performs better in terms of AP. Thus in what follows, without extra illustration, we adopt the naive subsampling as the default. 	
\begin{table}[tbh!]
\centering
\caption{Ablation results of downsampling operators in RRAM.}
\label{tab:DS}
\begin{tabular}{l|cccc}
	\hline
	 & DS & AP & $\text{AP}_{50}$ & $\text{AP}_{75}$ \\ 
	\hline
	baseline & - & 30.8& 53.8 & 31.7 \\
	~~~+RRAM & GAP & 31.6 & 55.7 & 32.2 \\
	~~~+RRAM & max pooling & 31.6  & 55.9 &  \textbf{33.6} \\
	~~~+RRAM & linear projection & 31.1 & 55.0 & 32.6 \\
	~~~+RRAM & naive subsample & \textbf{32.0} &  \textbf{56.1} & 32.6 \\		
	\hline
\end{tabular}
\end{table}

\textbf{Channel compression in RRAM.} We also investigate the influence of different settings of the channel number $C'$ in RRAM. Table \ref{tab:channels} lists the results when $C'$ is set to 64, 128 and 256 separately, which shows that our mehtod is robust to the choice of $C'$, and when $C'=128$, the proposed RRAM achieves the best performance.	
\begin{table}[tbh!]
\centering
\caption{Influences of different settings of $C'$ in RRAM.}
\label{tab:channels}
\begin{tabular}{c|ccc}
	\hline		
	$C'$ & AP & $\text{AP}_{50}$ & $\text{AP}_{75}$ \\ 
	\hline
	64 & \textbf{32.0}  & 56.0 & \textbf{33.0} \\
	128 & \textbf{32.0} & \textbf{56.1} & 32.6 \\
	256 & 31.7  & 55.8 & 32.4 \\		
	\hline
\end{tabular}
\end{table}

\textbf{Global feature map selection in GRAM.} Our GRAM reuses the feature maps from FPN \cite{lin2017feature} as global context for RoI feature enhancement. Usually, FPN has four levels and therefore outputs four scales of feature maps. From bottom to top, their resolution is $\frac{H}{2^{r+1}} \times \frac{W}{2^{r+1}}$ where $r\in [1, 4]$ is the level index. 
Therefore we also conduct experiments to determine which level of feature maps work better. Note that the resolution of feature maps at the bottom layer is too high to feed into GRAM with limited GPU memory, we downsample it with ratio of 4. The ablation results are listed in Table \ref{tab:ablation_results_RoI_extractor}, from which we have the following observations. First, enhancing RoI feature with the feature map from any level of FPN do bring performance gains. Second, utilizing the bottom level of feature maps in FPN achieves better performance than the top level. Third, using the bottom level feature map requires a larger interval of downsampling, but it can still get better performance. Based on above observations, the proposed GRAM explores global context from the first-level FPN feature maps.
\begin{table}
\centering
\caption{Ablation results of FPN feature selection in GRAM.}
\label{tab:ablation_results_RoI_extractor}
\begin{tabular}{l|cccc}
	\hline
	 & FPN level & AP & $\text{AP}_{50}$ & $\text{AP}_{75}$  \\ \hline
	baseline & - & 30.8 & 53.8 & 31.7 \\
	~~~+GRAM & level 1 & \textbf{31.8} & \textbf{56.2} & \textbf{32.7} \\
	~~~+GRAM & level 2 & 31.7 & \textbf{56.2} & 32.3 \\
	~~~+GRAM & level 3 & 31.5 & 55.8 & 32.2 \\
	~~~+GRAM & level 4 & 31.6 & 55.8 & 32.3 \\
	\hline
\end{tabular}
\end{table}

\textbf{Channel compression in GRAM.} We also investigate the influence of different settings of the channel number $C''$ in GRAM. Table \ref{tab:channels_GRAM} reports results and shows that our GRAM achieves the best performance when $C''$ is set to 128.
\begin{table}
\centering
\caption{Influences of different settings of $C''$ in GRAM.}
\label{tab:channels_GRAM}
\begin{tabular}{c|ccc}
	\hline		
	$C''$ & AP & $\text{AP}_{50}$ & $\text{AP}_{75}$ \\ 
	\hline
	64 & 31.6 & \textbf{56.2} & 32.3 \\
	128 & \textbf{31.8} & \textbf{56.2} & \textbf{32.7} \\
	256 & 31.5 & 55.7 & 32.3 \\		
	\hline
\end{tabular}
\end{table}

\begin{figure*}[tb!]
\centering
\includegraphics[width=.78\textwidth]{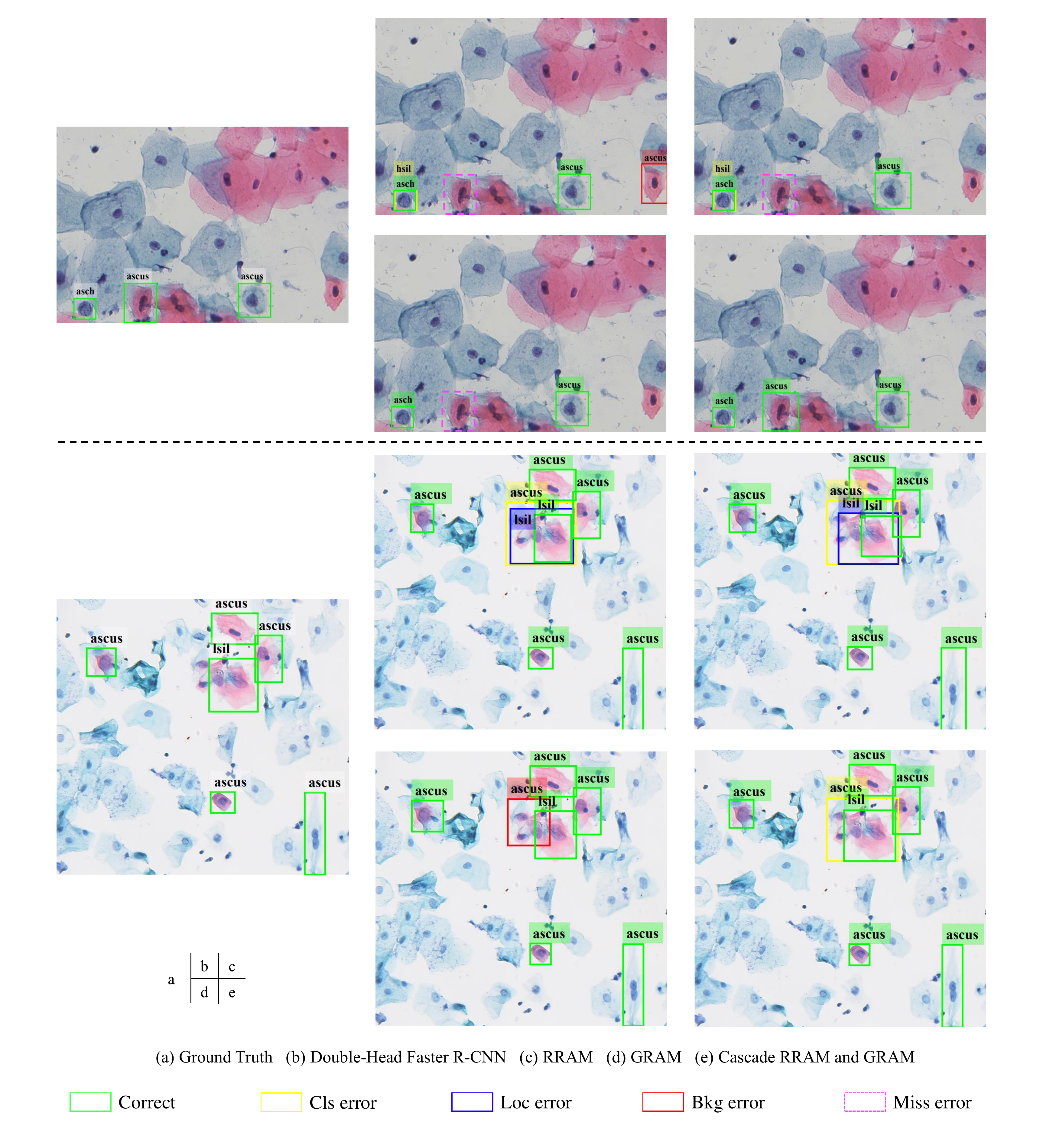}
\caption{Detection results visualization of two CCD test examples. Four detection errors (i.e., classification errors, location errors, background errors, and missing GT errors) are marked with different color. For the first example, the baseline fails to detect an `ASCUS' cell marked by dotted magenta box and triggers a missed GT error while cell detectors with our RRAM, Cascade RRAM and GRAM are able to correctly identify all `ASCUS' cells. Moreover, the baseline is confused by the background region in solid red box and makes background error while detectors with the proposed GRAM and Cascade RRAM and GRAM are able to avoid the background error. In the second example, the baseline encounters one classification error, one location error and one background error. Equipped with our RRAM, background error is fixed while with our Cascade RRAM and GRAM, the location error and background error are also fixed.}
\label{fig:error_visual}
\end{figure*}

\textbf{Combination strategies}. We investigate three strategies to combine the proposed RRAM and GRAM, i.e. parallelled attaching RRAM and GRAM on the top of RoI features and then fusing their outputs via element-wise summation, cascading RRAM and GRAM, and cascading GRAM and RRAM. The corresponding results are listed in Table \ref{tab:results_three_manners}. As it shown, comparing to the baseline, equipping with either RRAM or GRAM achieves better AP. Furthermore, combining RRAM and GRAM to enhance RoI feature can also boost the detection performance. Among the three combination strategies, Cascade RRAM and GRAM achieves the best detection result with an AP of 32.3.
\begin{table}
\centering
\caption{Results of combination strategies of GRAM and RRAM.}
\label{tab:results_three_manners}
\scalebox{0.9}
{\begin{tabular}{l|ccc}
		\hline
		 & AP  & $\text{AP}_{50}$ & $\text{AP}_{75}$ \\
		\hline
		baseline & 30.8 & 53.8 & 31.7 \\ 
		~~~+RRAM & 32.0 & 56.1 & 32.6 \\
		~~~+GRAM & 31.8 & 56.2 & 32.7 \\
		~~~+Parallel RRAM and GRAM & 32.2 & \textbf{56.6} & 33.0 \\
		~~~+Cascade GRAM and RRAM & 32.1  & 56.3 & 32.9 \\
		~~~+Cascade RRAM and GRAM & \textbf{32.3}  & \textbf{56.6} & \textbf{33.7} \\
		\hline
\end{tabular}}
\end{table}

\textbf{Architecture}. The proposed ROI feature enhancement modules are \textit{plug-and-play} that can be applied to any proposal-based object detector. We hereby integrate the cascade RRAM and GRAM methods into three commonly used methods, namely the original Faster R-CNN \cite{ren2017faster}, Double Head Faster R-CNN \cite{wu2020rethinking} and Cascade R-CNN \cite{cai2021cascade}. The results are shown in Table \ref{tab:results_different_detectors}, which show that the introduction of our module will consistently bring AP improvement. 
\begin{table*}[tb!]
\centering
\caption{Results of different two-stage object detectors with the proposed Cascade RRAM and GRAM.}
\label{tab:results_different_detectors}
\setlength\tabcolsep{4pt}
{\begin{tabular}{l|ccc}
		\hline
		 & AP & $\text{AP}_{50}$ & $\text{AP}_{75}$ \\
		\hline
		Faster R-CNN \cite{ren2017faster} & 30.5 & 54.0 & 31.3 \\
		~~+Cascade RRAM and GRAM & \textbf{31.6}(+1.1) & \textbf{56.1}(+2.1) & \textbf{32.3}(+1.0) \\ \hline
		Double-Head Faster R-CNN \cite{wu2020rethinking} & 30.8 & 53.8 & 31.7 \\ 
		~~+Cascade RRAM and GRAM & \textbf{32.3}(+1.5)  & \textbf{56.6}(+2.8) & \textbf{33.7}(+2.0) \\ \hline
		Cascade R-CNN \cite{cai2021cascade} & 31.1 & 53.2 & 32.8 \\
		~~+Cascade RRAM and GRAM & \textbf{31.6}(+0.5) & \textbf{53.9}(+0.7) & \textbf{33.6}(+0.8) \\ \hline	
\end{tabular}}
\end{table*}

\textbf{Error analysis}. We also perform error analysis to further provide insights of our methods using the recent proposed tool TIDE \cite{tide-eccv2020}, where the object detection errors are divided into six types (i.e., classification error, localization error, both classification and localization error, duplicate detection error, background error, missed ground-truth error), false positive, and false negative. Following \cite{tide-eccv2020}, we compute the $AP_{50}$ change (denoted as $E$) of each error and report the results in Table \ref{tab:error_analysis}. 
\begin{table*}[tb!]
\centering
\caption{Error analysis of different detection error types (Cls: classification error; Loc: localization error; Both: both classification and localization error; Dupe: duplicate predictions error; Bkg: background error; Miss: missing error; FP: false positive; FN: false negative).}
\label{tab:error_analysis}
\begin{tabular}{l|cccccccc}
	\hline
	methods & $E_{\mathrm{Cls}}$ & $E_{\mathrm{Loc}}$ & $E_{\mathrm{Both}}$ & $E_{\mathrm{Dupe}}$ & $E_{\mathrm{Bkg}}$ & $E_{\mathrm{Miss}}$  & $E_{\mathrm{FP}}$ & $E_{\mathrm{FN}}$ \\ \hline
	Double-Haed Faster R-CNN (baseline) & 5.29 & \textbf{5.76} & 0.50 & 0.11 & 15.81 & 2.76 & 34.16 & 6.80\\
	~~+RRAM & 4.97 & 6.28 & \textbf{0.48} & 0.12 & 14.76 & 2.62 & 32.46 & 6.68\\
	~~+GRAM & 5.18 & 6.39 & 0.50 & 0.11 & \textbf{14.58} & 2.92 & \textbf{31.53} & 7.25\\
	~~+Cascade RRAM and GRAM & \textbf{4.89} & 6.22 & \textbf{0.48} & \textbf{0.08} & 14.79 & \textbf{2.49} & 32.15 & \textbf{6.49} \\
	\hline
\end{tabular}
\end{table*}

As shown, compared to the baseline, the integration of both RRAM and GRAM as well as their combination can significant simultaneously reduce the classification error (Cls), and background error (Bkg), indicating that RRAM, GRAM as well as their combination are able to improve the discrimination ability of RoI features, leading to better performance on cell classification. 
Moreover, our method can decrease the number of false positives and false negatives. It should be noted that our methods produce larger location error, which is acceptable since for abnormal cell detection, the existence of abnormality is essential and therefore the correction classification arguably matters more than location \cite{tide-eccv2020}. 

\subsection{Comparison with SOTA methods} 
\begin{table*}[!htb]
	\centering
	\caption{Comparative results on CCD \texttt{test}. \dag means with multi-scale training.}
	\label{tab:efficient_table}
		\begin{tabular}{l|cccccccccc}
			\hline
			& AP & AP$_{50}$ & AP$_{75}$  & AP$_{S}$& AP$_{M}$& AP$_{L}$ & AR & Params & GFLOPs & FPS \\
			\hline
			Faster R-CNN \cite{ren2017faster} & 30.6 & 53.6 & 31.7 & 13.9 & 30.5 & 34.3 & 55.4 & 41.2 & 189.8 & 11.5 \\ 
			Double-Head Faster R-CNN \cite{wu2020rethinking} & 30.9 & 53.9 & 32.2 & 14.8 & 31.2 & 34.3 & 55.9 & 42.8 & 248.1 & 10.8\\ 
			Cascade R-CNN \cite{cai2021cascade} & 31.4 & 53.4 & 33.2 & 14.2 & 31.0 & 35.5 & 56.1 & 70.2 & 275.4 & 9.8 \\
			RetinaNet \cite{lin2018focal} & 29.9 & 52.5 & 31.1 & 11.8 & 31.0 & 33.8 & 57.3 & 36.3 & 192.8 & 13.1 \\
			FCOS \cite{tian2019fcos} & 29.6 & 52.3 & 30.5 & 11.4 & 30.2 & 33.4 & 55.6 & 31.9 & 182.4 & 19.1 \\		
			Non-local \cite{wang2018non} & 31.3 & 55.2  & 32.2 & 14.3 & 31.2 & 34.9 & 55.6 & 53.8 & 235.7 & 11.2\\
			RelationNet \cite{hu2018relation} & 23.7 & 46.5 & 21.8 & 8.8 & 23.7 & 27.9 & 47.9  & 45.4 & 211.9 & 10.8\\
			DETR\dag \cite{carion2020end} & 29.0 & 53.9 & 28.5 & 11.8 & 29.4 & 34.3 & 54.2  & 41.3 & 81.3 & 14.2\\
			deformble DETR\dag \cite{zhu2021deformable}  & 33.7 & 57.9 & 35.5 & 14.9 & 33.8 & \textbf{37.7} & \textbf{60.1}  & 39.8 & 195.2 & 8.7 \\
			Sparse R-CNN\dag \cite{sun2021sparse} & 31.7 & 54.4 & 22.6 & 14.9 & 32.1 & 34.1 & 58.2 & 105.9 &  149.9 & 9.9 \\	
			\hline
			YOLOv3* \cite{liang2021global} & 25.4 & 47.5 & 24.6 & 12.2 & 25.0 & 28.9 & 50.6  & 61.5 & 179.5 & 21.7\\
			AttFPN \cite{CAO2021102197} & 30.6 & 53.9 & 31.6 & 13.9 & 30.4 & 34.3 & 55.1  & 41.2 & 192.4 & 11.2\\ 
			TDCC-Net \cite{chen2022task} & 30.8 & 53.5 & 32.4 & 13.7& 30.8& 34.7& 55.8& 68.2  & 313.5 & 6.6 \\ 
			\hline
			RRAM & 32.0 & 56.0 & 32.8 & 15.7 & 31.8 & 35.6 & 55.9  & 42.9 & 302.2 & 10.3\\
			GRAM & 31.9 & 56.2 & 33.1 & 14.8 & 31.9 & 35.2 & 55.2 	& 42.9 & 261.1 & 10.4\\
			Cascade RRAM and GRAM & 32.4 & 56.6 & 33.5 & 15.5 & 32.8 & 35.3 & 56.3 & 43.1 & 311.4 & 9.2\\
			Cascade RRAM and GRAM\dag & \textbf{34.2} &\textbf{58.6} & \textbf{36.0} & \textbf{17.2} & \textbf{34.2} & 36.7 & 58.2  & 43.1 & 311.4 & 9.2\\
			\hline	
		\end{tabular}
\end{table*}

\begin{table*}[!htb]
	\centering
	\caption{Comparative per-class AP on the CCD \texttt{test}. \dag means with multi-scale training.}
	\label{tab:class_res_table}
		\begin{tabular}{l|cccccccccc}
			\hline
			& ASCUS & ASCH & LSIL & HSIL & AGC & VAG & MON & DYS & EC & AP \\
			\hline
			Faster R-CNN \cite{ren2017faster} & 29.9 & 22.2 & 30.9 & 33.2	& 42.4 & 28.8 &	19.1 & 50.3 & 18.7  & 30.6 \\ 
			Double-Head Faster R-CNN \cite{wu2020rethinking} & 30.4 & 22.7 & 31.2 & 33.3 & 42.8 & 29.1 & 18.4 & 51.1 & 19.2 & 30.9 \\
			Cascade R-CNN \cite{cai2021cascade} & 30.9 & 22.2 & 31.4 & 33.7 & 43.9 & 29.4 & 19.1 & 52.1 & 19.5 & 31.4 \\ 
			RetinaNet \cite{lin2018focal} & 31.4 & 20.6 & 32.0 & 32.2 & 43.4 & 24.0 & 16.9 & 51.2 & 17.6 & 29.9 \\
			FCOS \cite{tian2019fcos} & 31.2 & 20.8 & 32.2 & 31.5 & 42.3 & 23.5 & 16.2 & 52.1 & 16.2 & 29.6 \\		
			Non-local \cite{wang2018non} & 31.2 & 23.3 & 31.4 & 33.4 & 43.1 & 29.0 & 19.6 & 51.9 & 19.3 & 31.3\\
			RelationNet \cite{hu2018relation} & 21.2 & 16.2 & 26.1 & 27.9 & 35.4 & 20.0 & 14.2 & 41.1 & 11.5 & 21.8\\
			DETR\dag \cite{carion2020end} & 32.8 & 19.0 & 30.5 & 30.1 & 40.7 & 20.4 & 18.9 & 52.7 & 16.0  & 29.0\\
			deformable DETR\dag \cite{zhu2021deformable} & \textbf{36.9} & 24.2 & \textbf{34.8} & 34.3 & 45.3 & 27.1 & \textbf{23.7} & \textbf{56.0} & 20.9  & 33.7\\
			Sparse R-CNN\dag \cite{sun2021sparse} & 33.9 & 22.0 & 32.2 & 33.5 & 43.2 & 28.2 & 18.5 & 54.7 & 19.6 & 31.7 \\						
			\hline
			YOLOv3* \cite{liang2021global} & 29.0 & 18.2 & 27.2 & 29.3 & 37.6 & 20.9 & 8.7 & 40.5 & 17.4 & 25.4 \\
			AttFPN \cite{CAO2021102197} & 30.4 & 21.9 & 30.6 & 33.1 & 43.3 & 28.9 & 18.2 & 50.3 & 18.6 & 30.6 \\ 
			TDCC-Net \cite{chen2022task} & 30.4 & 22.2 & 30.5& 33.0 & 43.6 & 28.6 & 18.5& 52.1& 18.5& 30.8\\ 				
			\hline
			RRAM & 32.1 & 24.1 & 31.8 & 33.7 & 43.9 & 29.5 & 19.5 & 52.9 & 20.3 & 32.0 \\
			GRAM & 32.2 & 23.3 & 31.7 & 33.3 & 44.1 & 29.5 & 19.7 & 53.1 & 20.5 & 31.9 \\
			Cascade RRAM and GRAM & 32.2 & 24.1 & 32.9 & 34.0 & 44.4 & 29.7 & 20.6 & 53.4 & 20.7 & 32.4 \\
			Cascade RRAM and GRAM\dag & 35.2 & \textbf{25.3} & 34.5 & \textbf{35.6} & \textbf{46.2} & \textbf{29.8} & 22.0 &\textbf{ 56.0} & \textbf{22.8} & \textbf{34.2} \\
			\hline		
		\end{tabular}
\end{table*}

We compare our method to ten general object detectors and three cervical cell-specific detectors. The general object detectors includes Faster R-CNN \cite{ren2017faster} with FPN \cite{lin2017feature}, Double-Head Faster R-CNN \cite{wu2020rethinking}, Cascade R-CNN \cite{cai2021cascade}, RetinaNet \cite{lin2018focal}, FCOS \cite{tian2019fcos}, Non-local \cite{wang2018non}, RelationNet \cite{hu2018relation}, DETR \cite{carion2020end}, Deformable DETR \cite{zhu2021deformable} and Sparse R-CNN \cite{sun2021sparse}. 
The cervical cell detectors are YOLOv3* \cite{liang2021global}, AttFPN \cite{CAO2021102197} and TDCC-Net \cite{chen2022task}. All of those methods adopt ResNet-50 as backbone except for YOLOv3* which adopts Darknet-53 \cite{redmon2018yolov3}. We train most models for 12 epochs, except for YOLOv3*, DETR, deformable DETR and Sparse R-CNN. We train YOLOv3* for 30 epochs, DETR for 150 epochs, and deformable DETR and Sparse R-CNN for 50 epochs. As suggested, we apply the standard multi-scale training trick for DETR \cite{carion2020end}, deformable DETR \cite{zhu2021deformable}, Sparse R-CNN \cite{sun2021sparse}. 

Table \ref{tab:efficient_table} reports the comparative detection performance, along with the parameters, GFLOPs and FPS of different detectors. As we can see, cell detectors equipped with either our RRAM or GRAM exceeds the baseline (i.e. Double-Head Faster R-CNN \cite{wu2020rethinking}), by 1.1 and 1.0 AP, respectively. After cascading GRAM and RRAM, the performance is further boosted, achieving a 32.4 AP and surpassing the baseline by 1.5 AP, with the introduction of only extra 0.3M parameters which are almost negligible. Since attention mechanism involves a large number of matrix operations, GFLOPs of our method increases a lot but the FPS is still comparable to the baseline. When adding a multi-scale training strategy as it was done in Deformable DETR \cite{zhu2021deformable}, our method achieves 34.2 AP, which significantly exceeds all general object detectors and specialized cervical cell detectors. 

As shown in Table \ref{tab:efficient_table}, one-stage methods, i.e., RetinaNet \cite{lin2018focal}, FCOS \cite{tian2019fcos} and YOLOv3* \cite{liang2021global}, usually achieve higher FPS and therefore faster inference speed but much lower AP. While compared to attention-based detectors, i.e., Non-local \cite{wang2018non}, RelationNet \cite{hu2018relation}, DETR \cite{carion2020end} and Deformable DETR \cite{zhu2021deformable}, the proposed method achieves superior AP with competitive inference speed. It is worthwhile to mention that our methods have fewer parameters than Non-local \cite{wang2018non} and only need less than one-tenth of the training time of the DETR \cite{carion2020end}. Compared to deformable DETR \cite{zhu2021deformable} and Sparse R-CNN \cite{sun2021sparse}, our method also shows good performance after using the same multi-scale training strategy and training epochs. As for comparing to cervical cell specific detectors, i.e., YOLOv3* \cite{liang2021global}, AttFPN \cite{CAO2021102197}, and TDCC-Net \cite{chen2022task}, our method also achieves the highest AP. Although YOLOv3* \cite{liang2021global} has the fastest inference speed, its accuracy is limited by the input image resolution. Interestingly, both AttFPN \cite{CAO2021102197} and TDCC-Net \cite{chen2022task} are built on Faster R-CNN, but they perform similarly as the original Faster R-CNN on our CCD dataset. We also report the per-class AP in Table \ref{tab:class_res_table} which shows that our method obtains the best AP over 6 categories and comparative performance  over the rest categories, achieving substantially improvement. 

\section{Extended experiments: Image- and smear-level classification}
Our framework can easily be extended for downstream clinical cytopathological diagnosis applications such as FoV image-level classification and smear-level classification. The extension for image-level classification is conceptually simple: We make minor modifications by adding an extra image classification branch that output the class label of each FoV image \cite{liang2021global}. As for the extension for smear-level classification, we first aggregate the learned features of FoV images into a deep cytopathological feature and then feed it into a smear classifier to obtain the smear label \cite{geng2022}. It should be noted that the \textit{minimal} domain knowledge is adopted because the experiments are mainly to illustrate that our framework can facilitate both image- and smear-level classification. We expect specific designs will be complementary to our simple method, but it is beyond the scope of this paper.

\subsection{Image-level classification}	
\textbf{Implementation details.} Parallel to the detection head, we add a simple image-level classification head which consists of a convolutional layer, a GAP layer and a fully connected layer and is initialized with Kaiming method \cite{he2015delving}. Specifically, the bottom-level features of the FPN \cite{lin2017feature} are fed into the image-level classification head. We adopt the standard cross-entropy loss and continue to train for 6 epochs jointly with image classification and cell detection and the loss of image classification head is weighted by 0.15. The learning rate is initialized by 0.0001 which decreases to 0.00001 after 3 epochs. We adopt SGD optimizer with momentum 0.9 and weight decay 0.0001.

\textbf{Dataset and evaluation metrics.} We use the same CCD dataset and the label of each image is deduced by the cell-level labels: If an image contains any abnormal instances (i.e., ASCUS, ASCH, LSIL, HSIL and AGC), this image is labeled as abnormal (ABN), otherwise it's label is Negative for Intraepithelial Lesion or Malignancy (NILM). The image category distribution is listed in Table \ref{tab:tct_image}. We adopt sensitivity (sens), specificity (spec) and precision (prec) as evaluation metrics \cite{liang2021global}.

\begin{table}[!htb]
\centering
\caption{The image-level class distribution in the CCD dataset. }
\label{tab:tct_image}
\begin{tabular}{l|cc}
	\hline
	& NILM & ABN \\ \hline
	\texttt{train}  & 11,215 & 18,785 \\
	\texttt{val} & 1,909 & 3,091 \\
	\texttt{test} & 1,858 & 3,142   \\
	\hline
\end{tabular}
\end{table}

\textbf{Results.}
We compare our method with the one which directly trains a ResNet \cite{he2016deep} for image classification, along with the method which adds a classification head on the Double-Head Faster R-CNN \cite{wu2020rethinking} for image classification. The comparative results of image-level classification on the CCD \texttt{test} set are shown in Table \ref{tab:tct_image_res}. Although all methods use the same ResNet-50 backbone, the jointly abnormal cells detection and image classification methods achieve better classification performances than the pure classification method. Furthermore, our method obtains the best performance in terms of all metrics, which indicates that the introduction of our cascade RRAM and GRAM is conducive to extract discriminant features which facilitating the image-level classification.

\begin{table}[!htb]
\centering
\caption{Comparison of image-level classification on CCD \texttt{test}. \texttt{ImgCls} means with the image-level classification head.}
\label{tab:tct_image_res}
\setlength\tabcolsep{4pt}
\begin{tabular}{l|ccc}
	\hline
	methods & sens & spec & prec \\ \hline
	ResNet-50  & 96.1 & 58.7 & 79.6  \\
	Double-Head Faster R-CNN + \texttt{ImgCls} & 97.0 & 90.9 & 94.7 \\
	Ours  + \texttt{ImgCls} & \textbf{97.2} & \textbf{92.1} & \textbf{95.4} \\
	\hline
\end{tabular}
\end{table}

\subsection{Smear-level classification}

\textbf{Implementation Details.} 
We adopt the leaning-based gigapixel WSI classification method \cite{geng2022} which first perform abnormal cells detection to learn FoV image-level representation and then aggregate them into a WSI-level feature for classification. 
We also make minor modifications by simply replacing the detection parts in \cite{geng2022} with the Double-Head Faster R-CNN equipted with our cascade RRAM and GRAM.

\textbf{Dataset and evaluation metrics.} We use the Cervical Smear Dataset (CSD) in \cite{geng2022}, which contains 2,625 cervical gigapixel WSIs, including 1,542 normal WSIs and 1,083 abnormal WSIs. We train using the CSD \texttt{train} set and report results on CSD \texttt{test} set. Following \cite{lin2021dual,geng2022}, we adopt sensitivity (sens), specificity (spec), precision (prec), sens.C (the sensitivity exclude ASCUS and ASCH ) and sens.H (the sensitivity of high risk types including ASCH, HSIL and SCC) as evaluation metrics.

\textbf{Results.} We compare our method with two SOTA methods \cite{lin2021dual,geng2022}, along with the traditional methods, and the results are listed in Table \ref{tab:tct_smear}. As it illustrated, our method achieves substantial improvements in both specificity and precision while maintaining comparative sensitivity. Particularly, our method obtains higher sensitivity of high risk types and of abnormal types excluding atypical ones.

\begin{table}[!htb]
\centering
\caption{Comparison of smear-level classification on CSD \texttt{test} set. Sens.C means the sensitivity exclude ASCUS and ASCH  and sens.H means the sensitivity of high risk types including ASCH, HSIL and SCC. 0.90 and 0.95 mean two constraints of ``minimum-required'' sensitivity \cite{lin2021dual}.}
\label{tab:tct_smear}
\setlength\tabcolsep{3pt}
\begin{tabular}{l|ccccc}
	\hline
	methods & sens & spec & prec & sens.C & sens.H \\ \hline
	Adaboost \cite{geng2022} & 73.8 & 87.7 & 80.7 & 82.9 & 87.5 \\
	RandForest \cite{geng2022} & 65.4 & 91.8 & 84.8 & 75.6 & 83.3 \\
	SVM \cite{geng2022} & 79.9 & 90.0 & 84.9 & 89.0 & 91.7 \\
	Lin-0.90 \cite{lin2021dual} & 92.3 & 30.7 & 48.3 & 95.9 & 95.8 \\
	Lin-0.95 \cite{lin2021dual} & 95.7 & 27.7 & 48.1 & 97.6 & 96.3 \\
	Geng-0.90 \cite{geng2022} & 94.4 & 90.9 & 87.9 & 96.3 & 95.8 \\
	Geng-0.95 \cite{geng2022} & \textbf{97.8} & 85.5 & 82.6 & 98.8 & \textbf{100} \\
	\hline
	Ours-0.90 & 94.4 & \textbf{97.6} & \textbf{96.5} & 98.8 & \textbf{100} \\
	Ours-0.95 & \textbf{97.8} & 86.8 & 83.9 & \textbf{100} & \textbf{100} \\
	\hline
\end{tabular}
\end{table}	
\section{Conclusion}
In this paper, we propose two universal modules, namely RRAM and GRAM, to exploit both contextual relationships between cells and cell-to-global images to enhance the RoI features representation. Particularly, RRAM and GRAM use the attention mechanism to enrich the RoI features with cells relationships and global context respectively, which can be easily integrated into the proposal-based detectors. Based on the proposed RRAM and GRAM modules, we propose a novel cascaded RoI feature enhancement scheme to boost the performance of abnormal cervical cell detection. Our results on a large-scale cervical cell detection dataset consisting of 40,000 cytology images show that our method can outperform previous SOTA general object detectors and cervical cell-specific detectors with significant margins. We also show that our feature enhancing scheme can facilitate the downstream clinical tasks including FoV image-level classification and smear-level classification.

\bibliographystyle{IEEEtrans}
\bibliography{refs.bib}

\end{document}